\documentclass[journal]{IEEEtran}
\ifCLASSINFOpdf
\else
\fi
\usepackage{makecell}
\usepackage{multirow}
\usepackage{graphicx}
\usepackage{caption}
\begin{document}
%
\title{PILAE: A Non-gradient Descent Learning Scheme for Deep Feedforward Neural Networks$^*$}
%
%
%

\author{Ping~Guo,~\IEEEmembership{Senior~Member,~IEEE,}
        Ke~Wang,
        and~XiuLing~Zhou
\thanks{The research work described in this paper was supported  by the grants from the National Key Research and Development Program of China (No. 2018AAA0100203),  Beijing Natural Science Foundation under Grant (No. 4162027), China Postdoctoral Science Foundation (2020M682348), and the Key Research Foundation of Henan Higher Education Institutions (21A520002).}
\thanks{P. Guo is with the Image Processing \&\ Pattern Recognition Lab, School of Systems Science, Beijing Normal University, Beijing, China, 100875 (e-mail: pguo@bnu.edu.cn). He is the author to whom the correspondence should be addressed.}
\thanks{K. Wang, co-first author, is with the School of Information Engineering, Zhengzhou University, Zhengzhou, China, 450001 (e-mail: iekwang@zzu.edu.cn). }
\thanks{X. L. Zhou, co-first author, is with the Department of Technology and Industry Development, Beijing City University, Beijing, China, 100083 (e-mail: zxlmouse@bcu.edu.cn).}
\thanks{* This work has been submitted to the IEEE for possible publication. Copyright may be transferred without notice, after which this version may no longer be accessible.}}

%
%

\markboth{This draft version has been submitted to the IEEE Trans. }%
{Shell \MakeLowercase{\textit{et al.}}: Bare Demo of IEEEtran.cls for IEEE Journals}

%



\maketitle

\begin{abstract}
In this work, a non-gradient descent learning (NGDL) scheme was proposed for deep feedforward neural networks (DNN). It is known that an autoencoder can be used as the building blocks of the multi-layer perceptron (MLP) DNN, the MLP is taken as an example to illustrate the proposed scheme of pseudoinverse learning algorithm for autoencoder (PILAE) in this paper. The PILAE with low rank approximation is a NGDL algorithm, and the encoder weight matrix is set to be the low rank approximation of the pseudoinverse of the input matrix, while the decoder weight matrix is calculated by the pseudoinverse learning algorithm. It is worth to note that only very few network structure hyper-parameters need to be tuned compared with classical gradient descent learning algorithm. Hence, the proposed algorithm could be regarded as a quasi-automated training algorithm which could be utilized in automated machine learning field. The experimental results show that the proposed learning scheme for DNN could achieve better performance on considering the tradeoff between training efficiency and classification accuracy. 
\end{abstract}


%
\IEEEpeerreviewmaketitle

\section{Introduction}

\IEEEPARstart{R}{epresentation} learning is to make the given learning task easier by learning useful information and extracting essential features from the data \cite{IEEEhowto:Bengio2013}. As a way of hierarchical representation learning, deep learning has achieved state-of-the-art performance in a broad range of domains, such as image and video recognition \cite{IEEEhowto:Kamel, IEEEhowto:Muhammad, IEEEhowto:Yaseen}, fault diagnosis \cite{IEEEhowto:Liu2019}, electrocardiogram (ECG) signal analysis \cite{IEEEhowto:Pourbabaee} and speech recognition \cite{IEEEhowto:Le}. There are two paradigms developed parallel in deep learning: one roots in probabilistic graphical models and the other roots in neural networks \cite{IEEEhowto:Bengio2013, IEEEhowto:Kamyshanska}. In literature, both of them focus on single layer greedy learning modules and their variants. Among the building block modules, two main modules are explored intensively. One is the restricted Boltzmann Machine (RBM) that is the representation of the probabilistic model, while the network architecture is the same with multilayer neural network in practical implementation \cite{IEEEhowto:Chen2015}. The other is the autoencoder that is the representation of the traditional multi-layer perceptron (MLP).

Autoencoder \cite{IEEEhowto:Hinton} is a simple artificial neural network that utilizes the input as the output to learn the representation of the raw data in an unsupervised fashion. Thus, autoencoder can be viewed as an unsupervised feature learning model. Several variants of autoencoder are developed, such as sparse autoencoders for overcomplete representation \cite{IEEEhowto:Ranzato}, denoising autoencoders and contractive autoencoders for learning robust representation \cite{IEEEhowto:Vincent, IEEEhowto:Rifai}. Autoencoders can be used as the building blocks for constructing deep networks. During the construction process, deep neural network is trained in a greedy layer-wise scheme, and then the trained autoencoders (without the decoder) are stacked into a multi-layer neural model. In the stacked autoencoders, the output of the previous autoencoder is used as the input of the subsequent one. The output of different hidden layer of the network can be views as the different level features of the original data.

Most of the learning algorithms for training autoencoders are based on the gradient descent algorithm, such as the back propagation (BP) algorithm. The gradient descent-based algorithm (GDBA) is an iterative optimization algorithm which is time consuming. Hence, the main problem associated with stacked autoencoders based deep learning is the high computation cost, especially when the network structure is complex. Besides, most GDBAs require carefully tuning of several control parameters. These hyper-parameters, such as maximum epoch, step length and momentum, are crucial to the success of the algorithm. However, the hyper-parameters are usually related to specific application. Hence it is dependent on empirical tricks to select the appropriate hyper-parameters. Furthermore, the design of the network structure, i.e. the number of the hidden layers and the number of neurons in each hidden layer, is also a very difficult task for most users, and it is made by trial and error in general.

The connection weights in deep neural network are significantly redundant\cite{IEEEhowto:Denil}. Most of the weights can be predicted by using a small fraction of them. In order to improve the training efficiency, it is proposed to reduce the weights size during training by pruning any weight whose value approaches zero\cite{IEEEhowto:Yu2012}. Re-parameterization is an alternative way to reduce the weights size, which is the transformation of parameterization space of a model\cite{IEEEhowto:Amari}. Constraining the singular values of weight matrix \cite{IEEEhowto:Jia}, weight normalization \cite{IEEEhowto:Salimans} and factorizing the weight matrix by periodic functions \cite{IEEEhowto:Chandra} can accelerate the training speed to some extent. However, the iterative GDBA adopted in those methods dominates the time consuming.

From literature review, we know that the GDBAs either fail or suffer from significant difficulties for some types of simple problems in deep learning \cite{IEEEhowto:Shwartz}. It is necessary to develop non-gradient based learning algorithms for deep learning. In this work, a fast and quasi-automated learning algorithm, that is, pseudoinverse learning algorithm with low rank approximation is proposed for training the autoencoder which will be used as the building blocks of the deep neural network. The proposed algorithm is based on the pseudoinverse learning algorithm which can train the feedforward network in an analytical way \cite{IEEEhowto:Guo1995, IEEEhowto:Guo2001,IEEEhowto:Guo2004}. During the training process, the rank and the low rank approximation \cite{IEEEhowto:Markovsky} of the input matrix are obtained by singular value decomposition (SVD) technique. The rank of the input matrix can be used to guide the setting of the number of hidden neurons. The encoder weight matrix is set to be the low rank approximation of the pseudoinverse of the input matrix. The decoder weight matrix is calculated by the pseudoinverse learning algorithm. 

In addition to the learning algorithm, the model architecture design is another important issue to develop a high-quality machine learning application. However, this task is not only time-consuming but always requires lots of experienced human experts. In order to make machine learning easier to apply or even take the human experts out of the learning process completely, a growing number of researches focus on automated machine learning (AutoML). For the deep learning, it is able to automatically learn features from the data, which could avoid the human intervention in the feature engineering. Therefore, the core issues are neural architecture search (NAS) and hyper-parameter optimization (HPO). NAS is a fundamental problem in AutoML. NAS has been successfully used to specify the model architecture for different learning tasks. Currently, NAS techniques mainly include evolutionary methods \cite{IEEEhowto:Liang, IEEEhowto:Liu2018}, reinforcement learning \cite{IEEEhowto:Zoph2017}, random search \cite{IEEEhowto:Li2019}, gradient-based methods \cite{IEEEhowto:Liu2019DARTS}, and Bayesian optimization \cite{IEEEhowto:Kandasamy}. NAS is usually followed by HPO which is to tune the hyper-parameters of the model in order to further optimize the performance. The HPO methods includes random search \cite{IEEEhowto:Bergstra}, Bayesian optimization \cite{IEEEhowto:Golovin, IEEEhowto:Mutny}, evolutionary computation-based method \cite{IEEEhowto:Chen2018}, particle swarm optimization-based method \cite{IEEEhowto:Lorenzo}, meta learning based method \cite{IEEEhowto:Maher}. Although existing NAS and HPO methods can achieve good performance on some learning tasks, they usually adopt trial and error strategy and require a large amount of computing resources and time. The proposed method in this work could be viewed as an alternative way towards AutoML with lower cost. The network in this work is able to grow dynamically layer by layer with early stop criterion. It is worth to note that there is no control hyper-parameters in training algorithm need to be tuned, and suggestion is given for the architecture hyper-parameter selection. Hence, the proposed algorithm can be regarded as a quasi-automated training algorithm for deep feedforward neural networks. 

Our main contributions are summarized as follows:

1) A fast non-gradient descent learning scheme is proposed for deep feedforward neural networks, which can overcome the shortcomings of the gradient descent based learning algorithms.

2) A quasi-automated learning scheme which can automatically determine the network architecture and avoid the control hyper-parameter tuning is presented.

\section{Related Works}
\subsection{\textit{Multi-layer Perceptron}}

Multi-layer perceptron (MLP) is a kind of feedforward neural networks, which was most studied in mid-eighties of last century. In our previous work \cite{IEEEhowto:Guo2003}, when error distribution is assumed as Gaussian type, under the framework of the Kullback–Leibler (KL) divergence, the error (loss) function has following form:
\begin{equation}\label{e1}
	J\approx{\frac{1}{2N\sigma^2}\sum_{i=1}^{N}\left(\parallel{\mathrm{z}_i-g(\mathrm{x}_i,\mathrm{W})}\parallel^2+h_\mathrm{x}\parallel{g'(\mathrm{x}_i,\mathrm{W})}\parallel^2\right)},
\end{equation}
where $g(\mathrm{x}_i,\mathrm{W})$ is a neural network mapping function, $\mathrm{x}_i$ is the input signal vector, $\mathrm{z}_i$ is the corresponding target output vector, and $h_\mathrm{x}$ is the smooth regularization parameter. When loss function with the form of Eq. (\ref{e1}) is applied to regularized autoencoder, it is the same with empirical loss in Ref. \cite{IEEEhowto:Alain}.

If we consider the general linear function mapping approximation in regularization term only, i.e. $g(\mathrm{x},\mathrm{W})=\mathrm{W}\mathrm{x}$, the regularization term becomes a weight decay form as follows:
\begin{equation}\label{e2}
	J\approx{\frac{1}{2N\sigma^2}\sum_{i=1}^{N}\parallel{\mathrm{z}_i-g(\mathrm{x}_i,\mathrm{W})}\parallel^2+\frac{1}{2\sigma^2}h_\mathrm{x}\sum_{j=1}^M{w_j^2}},
\end{equation}
where $M$ is the number of network weight parameters and $w_j$ is an element of the matrix $\mathrm{W}$ in a vector expression.

The regularization parameter $h_\mathrm{x}$ can be estimated based on training data as \cite{IEEEhowto:Guo2003}:
\begin{equation}\label{e3}
	h_\mathrm{x}\approx{d_\mathrm{x}^2\left[1+(d_\mathrm{x}-1)^2\right]\frac{\sum_{i=1}^N{\parallel{\mathrm{z}_i-g(\mathrm{x}_i,\mathrm{W})}\parallel^2}}{N\sum_{j=1}^M{w_j^2}}},
\end{equation}
where $d_\mathrm{x}$ is the dimension of input vector $\mathrm{x}$.

\subsection{\textit{Autoencoder}}

An autoencoder can be viewed as a simple feedforward neural network with one input layer, one hidden layer, and one output layer \cite{IEEEhowto:Hinton}. The goal of training an autoencoder is to learn data representation with minimal reconstruction error.

From the input layer to the hidden layer, the encoder represents the input vector $\mathrm{x}$ as a feature vector $\mathrm{y}$, which is realized by linear mapping and nonlinear activation function:
\begin{equation}\label{e4}
	\mathrm{y}=f\left(\mathrm{W}_\mathrm{e}\mathrm{x}\right).
\end{equation}

From the hidden layer to the output layer, the decoder reconstructs the input vector $\mathrm{x}$ as vector $\mathrm{z}$:
\begin{equation}\label{e5}
	\mathrm{z}=g_1\left(\mathrm{W}_\mathrm{d}\mathrm{y}\right).
\end{equation}

In Eq. (\ref{e4}) and Eq. (\ref{e5}),$\mathrm{W}_\mathrm{e}$ and $\mathrm{W}_\mathrm{d}$ are the connection weight matrices, $f$ and $g_1$ are activation functions. For concision, the bias parameters are omitted and $g_1$ is set to be linear function in this paper. If the bias neuron is considered, it is simple to expand weight matrix as an augmented matrix \cite{IEEEhowto:Bishop}. Weight parameters in an autoencoder are estimated by minimizing the average reconstruction error function:
\begin{equation}\label{e6}
	J=\frac{1}{N}\sum_{i=1}^{N}{\parallel{\mathrm{x}_i-\mathrm{W}_\mathrm{d}(f(\mathrm{W}_\mathrm{e}\mathrm{x}_i))}\parallel_2^2},
\end{equation}
where $N$ is the number of input data. Error function in Eq. (\ref{e6}) is the sum-of-square error function. If we study regularized autoencoder, empirical loss function in Eq. (\ref{e1}) should be considered.

The weights $\mathrm{W}_\mathrm{d}$ can be set as $\mathrm{W}_\mathrm{d}=\mathrm{W}_\mathrm{e}^T$, which is termed as tied weights so as to decrease the number of estimated free parameters in an autoencoder. Autoencoder can be seen as a representation transformation. When the number of hidden neurons is constrained to be less than the number of input neurons, a compressed data representation of the input data is obtained, which realizes the dimension reduction (feature extraction) task. The constraints of tied weights and hidden neuron number (information bottleneck) can prevent the autoencoder from learning identity mapping.

\subsection{\textit{Basic Pseudoinverse Learning Algorithm}}

Pseudoinverse Learning (PIL) Algorithm is a fast learning algorithm for feedforward neural networks proposed by Guo \emph{et al} \cite{IEEEhowto:Guo1995, IEEEhowto:Guo2001, IEEEhowto:Guo2004}. The basic idea of the PIL algorithm is to find a set of orthogonal vector bases and use the nonlinear activation function to make the output vector of the hidden layer neurons orthogonal, and then obtain the output weights of the network by calculating the pseudoinverse solution.

Suppose we have a given data set denoted as $D=\{\ {\mathrm{x}_i,\mathrm{o}_i} \}\ _{i=1}^N$, where $\mathrm{x}_i=(x_{i1},x_{i2},\dots,x_{id})^T\in R^d$ is the input vector and  $\mathrm{o}_i=(o_{i1},o_{i2},\dots,o_{id})^T\in R^m$ is the desired output vector. Denote $\mathrm{X}\in R^{d\times N}$ as the input matrix and $\mathrm{O}\in R^{m\times N}$ as the desired output matrix. (If input bias neuron is considered, input matrix $\mathrm{X}$ will be $\mathrm{X}\in R^{(d+1)\times N}$). Consider a feedforward neural network with $L$ layer, where the linear activation function is used in last layer. Let $\mathrm{W}^l$ be the weight matrix which connects the layer $l$ and the layer $l+1$. Denote $\mathrm{H}^l=f(\mathrm{W}^{l-1}\mathrm{H}^{l-1})$ as the output matrix of the layer $l$ where $f(\mathrm{x})$ is an activation function.

The task of training the network is trying to find the optimal weight matrices which minimize the error function, for example, the most used sum-of-square error function:
\begin{equation}
	\min\parallel{\mathrm{W}^L\mathrm{H}^L-\mathrm{O}}\parallel^2.
\end{equation}
Based on the generalized linear algebra theory \cite{IEEEhowto:Boullion}, for the equation $\mathrm{W}^L\mathrm{H}^L=\mathrm{O}$, the optimal approximation solution is $\mathrm{W}^L=\mathrm{O}(\mathrm{H}^L)^+$, where $(\mathrm{H}^L)^+$ represents the pseudoinverse of the matrix $\mathrm{H}^L$. Let $\mathrm{W}^L=\mathrm{O}(\mathrm{H}^L)^+$, the objective function becomes:
\begin{equation}
	\min\parallel{\mathrm{W}^L\mathrm{H}^L-\mathrm{O}}\parallel^2=\parallel{\mathrm{O}(\mathrm{H}^L)^+\mathrm{H}^L-\mathrm{O}}\parallel^2.
\end{equation}

For solving the above optimization problem, the connection weight $\mathrm{W}^L$ is set to be the pseudoinverse matrix $(\mathrm{H}^L)^+$, the matrix $\mathrm{H}^L$ is propagated feedforward and the rank of $\mathrm{H}^L$ is increased gradually by the nonlinear activation function. Once the rank of $\mathrm{H}^L$ is close to full, that is $(\mathrm{H}^L)^+\mathrm{H}^L$, is close to an identity matrix, namely, if $\parallel{(\mathrm{H}^L)^+\mathrm{H}^L-\mathrm{I}}\parallel^2<\epsilon$ the learning procedure can be terminated. The pseudoinverse matrix $(\mathrm{H}^L)^+$ can be considered as a set of orthogonal vector bases which project matrix $\mathrm{H}^L$ to a new hidden space.

The PIL algorithm is a feedforward algorithm without any iterative optimization process. Meanwhile, it is also an automated algorithm without critical user-specified parameters such as learning rate or momentum constant. Thus, PIL algorithm is much faster than the back propagation and other gradient descent-based algorithm. Furthermore, the structure of the network generated by PIL algorithm is the one with deep attribution, and the dynamical growth in depth is dependent on the given training data set.

\subsection{\textit{The Low Rank Approximation}}

Low rank matrix approximation is well studied in the numerical linear algebra community \cite{IEEEhowto:Markovsky}. Many classical matrix decomposition techniques are adopted for low rank approximation, such as singular value decomposition (SVD) and QR decomposition \cite{IEEEhowto:Loan}. In our proposed method, truncated SVD is used to calculate the rank of matrix and the low rank approximation of the pseudoinverse matrix.

For any matrix $\mathrm{A}\in R^{m\times n}$, it can be factorized by SVD as:
\begin{equation}
	\mathrm{A}=\mathrm{U}\Sigma\mathrm{V}^T,
\end{equation}
where $\mathrm{U}\in R^{m\times m}$ is an orthogonal matrix whose columns are the eigenvectors of $\mathrm{A}\mathrm{A}^T$, $\mathrm{V}\in R^{n\times n}$ is an orthogonal matrix whose columns are the eigenvectors of $\mathrm{A}^T\mathrm{A}$, and $\Sigma\in R^{m\times n}$ is an diagonal matrix whose entries are in descending order, $\sigma_1\geq\sigma_2\geq\dots\geq\sigma_r\geq 0$, along the main diagonal:
\begin{equation}
	\Sigma=diag\{\ \sigma_1,\dots,\sigma_r,0,\dots,0 \}\,
\end{equation}
with $r=\mathrm{rank}(\mathrm{A})$. In the above equation, $\sigma_1,\dots,\sigma_r$ are the square roots of the eigenvalues of $\mathrm{A}^T\mathrm{A}$, which are called the singular values of $\mathrm{A}$.

Mathematically, truncated SVD applied to a matrix will produce a low rank approximation to that matrix.

\section{Proposed Fast Learning Scheme For Multilayer Deep Neural Network}
For a given task with data set,  $D=\{\ {\mathrm{x}_i,\mathrm{z}_i} \}\ _{i=1}^N$, we assume that the mapping function $g(\mathrm{x},\mathrm{W})$ in Eq. (\ref{e1}) is a deep feedforward neural network (DNN), which has one input layer, one output layer and several hidden layers.
\begin{equation}\label{e11}
	g(\mathrm{x},\mathrm{W})=\mathrm{W}^{l}f(\mathrm{W}^{l-1}\cdots f(\mathrm{W}^{1}f(\mathrm{W}^{0}\mathrm{x}))\cdots).
\end{equation}

In our proposed learning scheme, ``divide and conquer'' strategy is utilized to train the DNN. With weight decay regularization loss function in Eq. (\ref{e2}), the formal solution of $\mathrm{W}^l$, which connects last hidden layer to output layer, is the pseudoinverse solution:
\begin{equation}\label{e12}
	\mathrm{W}^l=\mathrm{Z}\mathrm{Y}^T(\mathrm{Y}\mathrm{Y}^T+\lambda\mathrm{I})^{-1}=\mathrm{Z}\mathrm{Y}^+,
\end{equation}
where $\mathrm{Z}$ is target output matrix, $\mathrm{Y}$ is the last hidden layer output matrix,
\begin{equation}\label{e13}
	\mathrm{Y}=f(\mathrm{W}^{l-1}\cdots f(\mathrm{W}^{1}f(\mathrm{W}^{0}\mathrm{x}))\cdots),
\end{equation}
and  $\mathrm{Y}^+$ has the form $\mathrm{Y}^+=\mathrm{Y}^T(\mathrm{Y}\mathrm{Y}^T+\lambda\mathrm{I})^{-1}$. This form is similar with limit form of Moore-Penrose inverse, but here $\lambda$ is the regularization parameter and it should not approach to zero. The weight matrices in Eq. (\ref{e13}) will be trained in an unsupervised layered-wise manner, that is, the network in Eq. (\ref{e13}) is taken as a stacked autoencoders as shown in Fig.1.  Following we will describe pseudoinverse learning algorithm for autoencoders.

\subsection{\textit{Proposed Pseudoinverse Learning Algorithm with Low Rank Approximation for Autoencoders}}

There are several PIL algorithm and variants \cite{IEEEhowto:Guo2019}, in this work we focus on PIL with low rank approximation scheme. For a data matrix $\mathrm{X}\in R^{d\times N}$, $\mathrm{X} = [\mathrm{x}_1,\mathrm{x}_2,\dots,\mathrm{x}_N]$, where $\mathrm{x}_i=(x_{i1},x_{i2},\dots,x_{id})^T\in R^d$ is the $i$-th sample with dimension $d$. For training autoencoder, the objective (loss) function (reconstruction error) can be written as:
\begin{equation}
	\min\parallel{\mathrm{W}_\mathrm{d}\mathrm{H}-\mathrm{X}}\parallel^2.
\end{equation}

The pseudoinverse approximate solution is used to solve this optimization problem:
\begin{equation}
	\mathrm{W}_\mathrm{d}=\mathrm{X}\mathrm{H}^+.
\end{equation}

According to the basic PIL algorithm, the connection weight $\mathrm{W}_\mathrm{e}$ should be set as $\mathrm{X}^+$, which means implicitly a constraint that the number of hidden neurons is equal to the number of input samples. In general, the number of input samples is far more than the dimension of the input data.

Here, autoencoder is used to learn data representation. In order to learn good representation of the data, constraints such as sparse and/or information bottleneck (IB) are most used.  The constraint of IB is that the number of hidden neurons should be less than the dimension of the input data. With either sparse or IB constraint, the identity mapping can be avoided, while data representation is learnt \cite{IEEEhowto:Bengio2013}. According to the manifold hypothesis that real world data presented in high dimensional spaces is likely to concentrate in the vicinity of non-linear sub-manifolds of much lower dimensionality \cite{IEEEhowto:Narayanan}, the number of hidden neurons should be equal or more than the rank of the input data matrix.

In order to learn the data representation, a pseudoinverse learning algorithm with low rank approximation for autoencoders is proposed in this paper. In the proposed method, the rank of the input matrix is used to guide the setting of the hidden neuron number. The encoder weight matrix is set to be the low rank approximation of the pseudoinverse of the input matrix, and the decoder weight matrix is calculated with PIL algorithm \cite{IEEEhowto:Wang2017}.

There are three merits for using low rank approximation with SVD. First, the number of hidden neurons can be set automatically with a formula according to the rank of input data matrix. Second, the encoder weight matrix can be calculated with truncated SVD. Third, those samples, which can be expressed as the linear combination of other samples, are merged. By merging the dependent samples by low rank approximation and encoding the input data by linear dimension reduction, the input data are mapped into a feature space, in which its intrinsic characteristics can be better reflected.

The implementation of the pseudoinverse learning algorithm with low rank approximation is described in detail as follows.

1) Determine the number of hidden neurons by a formula with the rank of input data matrix.

The rank of input matrix can be calculated by the SVD method. At first, the input data matrix $\mathrm{X}$ can be decomposed by the SVD as follows:
\begin{equation}
	\mathrm{X}=\mathrm{U}\Sigma\mathrm{V}^T,
\end{equation}
where $\mathrm{U}\in{R^{d\times d}}$ and $\mathrm{V}\in{R^{N\times N}}$ are orthogonal matrices, $\Sigma\in{R^{d\times N}}$ is a diagonal matrix where the diagonal elements are the singular values of $\mathrm{X}$. Then the rank of the input matrix $\mathrm{X}$ is obtained by counting the number of nonzero singular values in matrix $\Sigma$:
\begin{equation}
	r=\mathrm{rank}(\mathrm{X})=\mathrm{the\ number\ of\ nonzero}(\Sigma).
\end{equation}
For large scale data sets, the QR iteration algorithm could be adopted to calculate the pseudoinverse of the big matrix and then obtain the rank of the input matrix \cite{IEEEhowto:Yu2014}.

The number of hidden neurons is set according to the rank of input matrix $\mathrm{X}$. In order to avoid obtaining an identity mapping in the autoencoder, the number of hidden neurons should generally be less than the dimension of the input data. However, too less number of hidden neurons will lead to more loss of information during the process of mapping the input data to the hidden feature space, which will inevitably result in higher reconstruction error. Hence, the number of hidden neurons should not be too small. In our previous work \cite{IEEEhowto:Wang2017}, on considering the tradeoff between feature extraction and reconstruction error, the number of hidden neurons is assigned as 
\begin{equation}\label{e18}
	p=r+\alpha(d-r),\alpha\in [0,1],
\end{equation}
where $d$ is the dimension of the input data $\mathrm{x}$. In the case of the data matrix with \emph{i.i.d}. noise, whose rank is almost full, we can apply data preprocessing techniques to reduce the noise, or we can force the dimensionality of input data to reduce. Therefore, the number of hidden neurons can be set with a decay factor shown as follows:
\begin{equation}\label{e19}
	p=\beta d,\beta\in [0,1].
\end{equation}

 2) Calculate the encoder weight matrix as the low rank approximation of the pseudoinverse of the input matrix. The parameter $\alpha$ or $\beta$ is estimated based on reconstruction error.

From the result of the SVD of $\mathrm{X}$, the pseudoinverse of matrix $\mathrm{X}$ can be calculated as follows:
\begin{equation}
	\mathrm{X}^+=\mathrm{V}\Sigma^+\mathrm{U}^T,
\end{equation}
where $\Sigma^+\in R^{N\times d}$ is the diagonal matrix composed of the reciprocal of nonzero elements in matrix $\Sigma$.

Let
\begin{equation}
	\hat\mathrm{X}^+=\hat\mathrm{V}\Sigma^+\mathrm{U}^T\in R^{p\times d},
\end{equation}
be the $p$ low rank approximation of $\mathrm{X}^+$, where $\hat\mathrm{V}\in R^{p\times N}$ is the matrix composed of the first $p$ rows of $\mathrm{V}$. This approximation is similar with truncated SVD method, because the truncated SVD can be viewed as a kind of regularization \cite{IEEEhowto:Hansen}, hence it can serve as a dimension reduction technique. In mathematics, this also can be realized by defining a $p\times N$ diagonal matrix $\Omega$, in which those diagonal elements are 1 and non-diagonal elements are 0, and $\hat\mathrm{V}=\Omega\mathrm{V}$ \cite{IEEEhowto:Magdonismail}. The encoder weight matrix is set to be the low rank approximation of the pseudoinverse matrix:
\begin{equation}
	\mathrm{W}_\mathrm{e}=\hat\mathrm{X}^+.
\end{equation}

Here, the encoder matrix is calculated with the truncated SVD method. It should be noted that there is no randomness in the weight initialization process, which is different from the BP based algorithms. Then the input matrix $\mathrm{X}$ is mapped into the $p$-dimensional hidden feature space as
\begin{equation}
	\mathrm{H}=f(\mathrm{W}_\mathrm{e}\mathrm{X})\in R^{p\times N}.
\end{equation}
Here, $f(\cdot )$ is the activation function.


3) Calculate the decoder weight matrix with the PIL algorithm.

According to the basic idea of PIL algorithm, the optimal solution of equation of $\mathrm{W}_d\mathrm{H}=\mathrm{X}$ is the pseudoinverse solution $\mathrm{W}_d=\mathrm{X}\mathrm{H}^+$. Hence, the pseudoinverse of $\mathrm{H}$ needed to be calculated. It can also be calculated by using SVD or QR method. Alternatively, if $\mathrm{H}$ is a row full rank matrix, it can be calculated as
\begin{equation}
	\mathrm{H}^+=\mathrm{H}^T(\mathrm{H}\mathrm{H}^T)^{-1}.
\end{equation}

When $\mathrm{H}$ is a rank defective matrix. To avoid ill-posed problem, a weight decay regularization term is added to the objective function as in Eq. (\ref{e2}). This form is also called ridge regression in statistics, and when omitting some optimization irrelative constants, the weight decay regularization in the matrix form is shown as follows:
\begin{equation}
	J=\frac{1}{2}\parallel{\mathrm{W}_\mathrm{d}\mathrm{H}-\mathrm{X}}\parallel^2
	+\frac{\lambda_1}{2}\parallel{\mathrm{W}_\mathrm{d}}\parallel^2.
\end{equation}

The regularization solution to the optimization problem in above equation has  analytical solution. Taking derivative of $J$ to $\mathrm{W}_\mathrm{d}$ and let it be zero, it can be obtained
\begin{equation}
	(\mathrm{W}_\mathrm{d}\mathrm{H}-\mathrm{X})\mathrm{H}^T+\lambda_1\mathrm{W}_\mathrm{d}=0.
\end{equation}

Then the pseudoinverse of $\mathrm{H}$ is calculated as follows:
\begin{equation}\label{e27}
	\mathrm{H}^+=\mathrm{H}^T(\mathrm{H}\mathrm{H}^T+\lambda_1\mathrm{I})^{-1}.
\end{equation}
where $\lambda_1$ is a regularization parameter.

The decoder weights can be calculated as:
\begin{equation}
\mathrm{W}_\mathrm{d}=\mathrm{X}\mathrm{H}^T(\mathrm{H}\mathrm{H}^T
+\lambda_1\mathrm{I})^{-1}=\mathrm{X}\mathrm{H}^+.
\end{equation}

Here, the regularization parameter  $\lambda_1$ can be regarded as a smooth parameter as in Eq. (\ref{e3}), so it can be estimated with training data only. This regularization parameter controls the reconstruction error, here we set it to be a very small value to guarantee that matrix $(\mathrm{H}\mathrm{H}^T+\lambda_1\mathrm{I})$ is not singular.

After the decoder weight is obtained, the decoder weights and the encoder weights will be tied, that is, $\mathrm{W}_\mathrm{e}=\mathrm{W}_\mathrm{e}^T$. The tied weights can reduce the freedom degree of the model. The hidden feature matrix $\mathrm{H}=f(\mathrm{W}_\mathrm{e}\mathrm{X})$ can be used as the input data matrix for succeeding autoencoders. Note that the encoder weight matrix is tied with decoder weight hence  depends on. If it is changed, $\mathrm{H}$ will be changed also, resulting in  needs to be computed again. In theoretical analysis, it will diverge if weights are iterative computed. This problem might be solved by using SVD-free approximation technique we intend to develop later.

\subsection{\textit{Pseudoinverse Learning for Deep Stacked Autoencoders}}
Based on the proposed pseudoinverse learning algorithm with low rank approximation for autoencoders, a fast and quasi-automated learning algorithm for stacked autoencoders is developed, which can overcome the shortcomings of the gradient descent-based training algorithms for deep feedforward neural network such as BP algorithm.

In the proposed method, the pseudoinverse learning algorithm with low rank approximation is applied for training the autoencoder which will be used as the building blocks of the stacked autoencoders. After obtaining the encoder connection weight matrix $\mathrm{W}_\mathrm{e}$, the hidden output matrix $\mathrm{H}=f(\mathrm{W}_\mathrm{e}\mathrm{X})$ is used as the input matrix for succeeding autoencoders. This procedure is continued until a criterion is met. This criterion making the network stop growing may adopt a predesigned maximum number of layers or utilize the  $\parallel{\mathrm{H}^+\mathrm{H}-\mathrm{I}}\parallel^2<\epsilon$ as we did in previous work \cite{IEEEhowto:Guo2004}.

In fact, the proposed learning scheme can be explained as the greedy layer-wise learning strategy applied to the DNN, in which all trained autoencoders are then stacked to construct a deep neural network. In this work, the constructed network performs as a feature learning model instead of reconstructing the input data, hence the decoders of building blocks are all removed. However, in some cases, the deep network is expected to reconstruct the input data. In such cases, the decoders should be retained, and the outputs of the last hidden layer will be used to reconstruct the input from the previous layers to the first input layer in a layer-by-layer way. The For complex networks, the number of the weight parameters is reduced by tied weight operation, which means taking the transposed decoder weight matrix as the encoder weight matrix. The structure of the deep neural network based on stacked pseudoinverse autoencoders is shown in Fig. 1, while the mathematical expression is in Eq. (\ref{e13}).

\begin{figure}[!t]
	\centering
	\includegraphics[width=1.0\columnwidth]{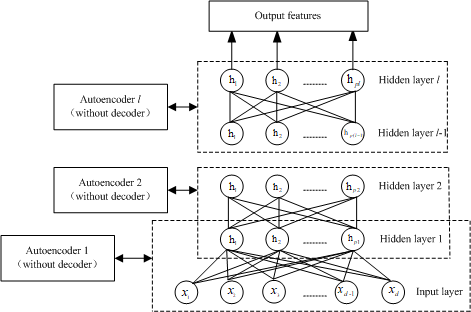}
	\captionsetup{justification=centering}
	\caption{Structure of the stacked pseudoinverse learning autoencoders deep network}
	\label{fig_1}
\end{figure}

After training, the output feature matrix of the stacked autoencoders based deep neural network can be calculated as Eq. (\ref{e13}), and the weight matrices are
\begin{equation}
	\mathrm{W}^i=\mathrm{W}_\mathrm{e}^i, \quad i=0,1,\cdots,l-1.
\end{equation}

With learnt feature matrix, we can use Eq. (\ref{e12}) to calculate the output weight matrix. If users intend to use the features for classification task and do not utilize single hidden layer neural net (SHLN) as we did, an additional classifier should be used, such as Softmax, support vector machine or other classifiers, to obtain final results.

In order to realize global optimization for the DNN structure, in practice, we need to slightly adjust the neuron number $p_\mathrm{L}$ and output layer regularization parameter $\lambda$ of the last hidden layer to minimize generalization error. Here we propose an empirical formula to estimate $p_\mathrm{L}$ based on the rank of input data and the number of training samples:
\begin{equation}\label{e30}
	p_\mathrm{L}=\left\{
	\begin{array}{rcl}
		P(r,N) & & { \mathrm{if} \quad P(r,N) > 0}\\
		r+\alpha (d-r) & & { \mathrm{otherwise}}
	\end{array} \right.
\end{equation}
where
\begin{equation}\label{e31}
	P(r,N)=\lceil{\theta_0+\theta_1 r+\theta_2 N+\theta_3 r^2+\theta_4 N^2}\rceil
\end{equation}
and $r$ is the rank of the input data, $N$ represents the number of training samples, $\theta_i(i=1,2,3,4)$ are constant coefficients.

\section{Experiment Results}
In this section, we first conduct experiment 1 to demonstrate how to determine the coefficients in Eq. (\ref{e31}). Then experiment 2 is conducted to compare our proposed method with the baseline. Since the BP algorithm is the representative gradient descent-based algorithm for training the autoencoders, the proposed algorithm is compared with the BP algorithm on different data sets to evaluate its performance. The deep neural network is trained in the classical greedy layer-wise scheme by the two algorithms respectively. Each hidden layer is associated with a trained autoencoder (without the decoder). The output of the last hidden layer can be used as the input features for specific learning tasks, e.g. classification.

\subsection{\textit{Experiments Setup}}
In the first experiment, we train SHLNs with several publicly available data sets. These data sets are different in terms of feature number, rank, and sample number. For each data set, the relationship between the hidden neural number and the classification performance is analyzed. Specifically, a series of SHLNs with different hidden neuron number are trained on each data set so as to reliably find a well-performed network structure which can achieve a relatively high accuracy. 5-fold cross validation is employed for evaluation of all networks. With the obtained well-performed network structure, regression analysis is then used to find the relationship between the rank of data along with the sample number and the optimal hidden neuron number, i.e. the equation defined in Eq. (\ref{e30}).

In the second experiment, the proposed algorithm and the BP algorithm are used to train networks with the same structure, respectively. A classification task is performed after the representation learning. To be specific, the output of the stacked autoencoders is taken as the input of the classifiers including the Softmax and SHLN. 
For the SHLN classifier, the weights of the last layer can be calculated with Eq. (\ref{e12}). $g(\mathrm{x},\mathrm{W})$ expressed in Eq. (\ref{e11}) is used for classification task. 

In the comparison experiment, the classifiers share the same parameters in both algorithms. Note that the main purpose of the experiments is not to pay attention to the classification accuracy, but to compare the performance of the feature learning capacity of different algorithms under the same conditions. Since the network structure can be quasi-automatically generate proposed learning scheme, we do not have to specify the network structure manually. To conduct a fair comparison, the same network structure is employed by both our algorithm and the BP algorithm. We use Eq. (\ref{e19}) to set the network architecture since the data sets are almost full rank. Specifically, setting $\beta$ to 0.9 derives the network architecture of 784-705-634-10. For BP method, Adam optimizer was employed with a learning rate of 0.001, two exponential decay rates of 0.9, 0.999 respectively and a batch size of 128. The training time of the BP based method was measured when the validation accuracy is saturated. To this end, we set a proper maximum epoch by analyzing the learning curve. 

\subsection{\textit{Data Sets}}
The benchmark data sets used to analyze the relationship between the rank, sample number and the optimal hidden neuron number are from the OpenML community (http://openml.org) \cite{IEEEhowto:Vanschoren} and UCI repository \cite{IEEEhowto:Lichman}. The data sets are described in TABLE I. Besides, MNIST, Fashion-MNIST, NORB and SVHN are used to compare the performance of the baseline methods and the proposed one. MNIST is a common data set which is usually used to evaluate the performance of the deep learning model. MNIST consists of 70,000 handwritten digit images with 28×28 pixels in gray scale. The digit images belong to 10 categories. Compared with MNIST, Fashion-MNIST \cite{IEEEhowto:Xiao} poses a more challenging classification task. It consists of a training set of 60,000 samples and a test set of 10,000 samples belonging to 10 categories. Each sample is a grayscale image of the same size with MNIST. NORB \cite{IEEEhowto:LeCun} has a total of 48,600 images of 3-D toy. It contains 50 toys belonging to 5 generic categories. SVHN \cite{IEEEhowto:Netzer} is a real-world image dataset which has over 600,000 digit images. Similarly with MNIST, SVHN has 10 distinct categories.

\begin{table}[!t]
	\renewcommand{\arraystretch}{1.5}
	\caption{The Summary Of The Data Sets}
	\label{table_1}
	\centering
	\begin{tabular}{llll}
		\hline
		Data set & \#Instance & \#Feature & Rank\\
		\hline
		Bupa & 250 & 6 & 6\\
		Segment & 1700 & 19 & 19\\
		Mfeat & 1400 & 649 & 646\\
		Prior & 2500 & 785 & 621\\
		Har 1 & 3000 & 562 & 540\\
		Advertisement & 3279 & 1558 & 727\\
		Gina\_agnostic & 3468 & 970 & 970\\
		Spambase & 4601 & 57 & 57\\
		Isolet & 7797 & 617 & 617\\
		Har 2 & 7352 & 562 & 540\\
		Sylva & 10000 & 216 & 204\\
		\hline
	\end{tabular}
\end{table}

\subsection{\textit{Results and Analysis}}
\textit{C.1 Results of experiment 1}

To verify the obtained relationship between the rank, the sample number and the optimal hidden neuron number, a leave-one-out cross validation is employed. To be specific, we first leave aside a single data set as a validation set, and use other data sets to generate the relationship specified in Eq. (\ref{e31}) by bivariate regression analysis. Then we use the relationship to predict the hidden neuron number and train a SHLN on the validation set. The classification accuracy of this SHLN with different learning algorithms is then compared with the well-performed one which is obtained by brute force search with a series of different hidden neuron numbers. The network obtained by brute search is trained with our PIL algorithm. The classification performance comparison between the predicted network structure with different learning algorithm and the well-performed structure obtained by search is summarized in TABLE II, in which the columns of ``\#Hidden neurons*'' and ``Test accuracy*'' denote the architectures and classification performance of the network structures determined with the brute force search strategy while ``Test accuracy$^\dag$'' and ``Test accuracy'' represents the results obtained by using BP algorithm and the proposed learning algorithm respectively with exactly same network architecture specified by Eq. (\ref{e31}). In addition, the training time of the proposed PIL algorithm and the BP based algorithm were also compared in this experiment and the results were shown in the last two columns of TABLE II. Column ``Training time'' represents the elapsed training time of the proposed scheme while the ``Training time$^\dag$'' denotes the one of BP based learning algorithm. It can be observed that the predicted network structure can achieve comparable accuracy to others except the first two data sets whose \#features are extremely low. However, the proposed scheme would take much less training time than BP based learning algorithm. Besides, as is shown in TABLE I, the data sets vary significantly in  \#features and \#instances, hence, the data sets could be categorized according to their rank (or \#features) and \#instances, and then conduct the regression analysis in each category respectively. For example, all data sets with low \#features, but large \#instances should be analyzed together while the ones with high \#features and medium \#instances are analyzed together.  

\begin{table*}[!t]
	\renewcommand{\arraystretch}{3}
	\caption{The Comparison Between The Accurcy Of The Predicted Network Strcture And The Optimal Structure With Trining Algorithms}
	\label{table_2}
	\centering
	\begin{tabular}{lllllllll}
		\hline
		Data set &  $[\theta_0,\theta_1,\theta_2,\theta_3,\theta_4]$  &  \makecell{\#Hidden \\ neurons} & \makecell{\#Hidden \\ neurons*} & \makecell{Test \\ accuracy} & \makecell{Test \\ accuracy\dag} & \makecell{Test \\ accuracy*} & \makecell{Training \\ time (sec)} & \makecell{Training \\time\dag (sec)}\\
		\hline
		Bupa & \makecell[l]{[-1.2982e3,5.0603,1.0390,\\-0.0036,-9.0308e-5]} & -1064 & 161 & - & - & 81.16 & - & - \\
		Segment & \makecell[l]{[-926.2121,5.6153,0.6169,\\-0.0037,-4.1785e-5]} & -52 & 1352 & - & - & 95.67 & - & - \\
		Mfeat & \makecell[l]{[-377.4237,4.7423,0.6002,\\-0.0033-4.5829e-05]} & 1921 & 1265 & 98.75 & 97.50 & 98.75 & 0.60 & 47.35 \\
		Prior & \makecell[l]{[-434.6510,5.2149,0.6614,\\-0.0039,-5.4856e-5]} & 2416 & 1979 & 89.32 & 91.20 & 90.19 & 1.23 & 59.33 \\
		Har 1 & \makecell[l]{[-515.3591,5.5879,0.7571,\\-0.0045,-6.714e-5]} & 2627 & 1557 & 97.67 & 97.33 & 97.83 & 0.99 & 146.90 \\
		Advertisement & \makecell[l]{[-459.4272,4.2053,0.6779,\\-0.0029,-5.2552e-5]} & 2497 & 2556 & 95.42 & 97.56 & 95.42 & 2.56 & 171.79 \\
		Gina\_agnostic & \makecell[l]{[-373.4045,-1.9230,0.7241, \\0.0063,-4.7683e-5]} & 5339 & 2197 & 84.70 & 86.29 & 84.99 & 5.14 & 237.48 \\
		Spambase & \makecell[l]{[-712.0285,0.7562,1.8134,\\-0.0013,-0.0002]} & 3510 & 586 & 91.09 & 92.80 & 90.65 & 1.70 & 24.57 \\
		Isolet & \makecell[l]{[-293.6718,3.8698,0.6184,\\-0.0026,-5.1742e-5]} & 2949 & 4062 & 92.69 & 47.27 & 93.39 & 2.49 & 210.99 \\
		Har 2& \makecell[l]{[-58.6140,3.0542,0.4126,\\-0.0014,-2.9070e-5]} & 2603 & 4657 & 98.03 & 98.16 & 99.05 & 1.82 & 289.51 \\
		Sylva & \makecell[l]{[487.8985,3.04204,-0.2173,\\-0.0014,0.0001]} & 5893 & 1322 & 98.61 & 94.03 & 98.85 & 9.41 & 34.51 \\
		
		\hline
	\end{tabular}
\end{table*}

\textit{C.2 Results of experiment 2}

The comparison of the accuracy and elapsed training time on MNIST and Fashion-MNIST is shown in TABLE III and Fig. 2 respectively. For BP based learning algorithm, the average classification accuracy with the standard deviation were obtained by repeating the experiment fifty times. Note that there is no randomness in the optimization process of the proposed method, hence, no statistics were provided. Considering that exactly same classifiers are employed, hence the classification results mean that the neural network trained by our proposed algorithm has comparable representation learning capacity. In Fig. 2, the logarithm base 10 of the elapsed training time is used to scale the vertical coordinate. It can be observed that the proposed algorithm is much faster than the BP algorithm on both datasets. This is due to that the proposed algorithm is a non-gradient method without iterative optimization, whereas the BP based algorithm is a gradient based method which requires iterative optimization. Especially when the data size is large and/or the network is deep, the training with BP algorithm is extremely time-consuming. 

\begin{table}[!t]
	\renewcommand{\arraystretch}{3}
	\caption{THE Comparisiona Of Test Accuracy With Differnent Algorithms}
	\label{table_3}
	\centering
	\begin{tabular}{ccccc}
		\hline
		\multirow{2}{*}{} &
		\multicolumn{2}{c}{BP algorithm} &
		\multicolumn{2}{c}{Our proposed} \\
		\cline{2-5}
		& Softmax & SHLN & Softmax & SHLN \\
		\hline
		MNIST & \makecell{0.9771} & \makecell{0.9727} & 0.9611 & 0.9751 \\
		\hline
		Fashion-MNIST & \makecell{0.8843} & \makecell{0.8795} & 0.8621 & 0.8819 \\ 
		\hline
		NORB & \makecell{0.8723} & \makecell{0.8676} & 0.8913 & 0.9023 \\ 
		\hline
		SVHN & \makecell{0.7762} & \makecell{0.7807} & 0.7716 & 0.7785 \\ 
		\hline
	\end{tabular}
\end{table}

\begin{figure}[!t]
	\centering
	\includegraphics[width=1.0\columnwidth]{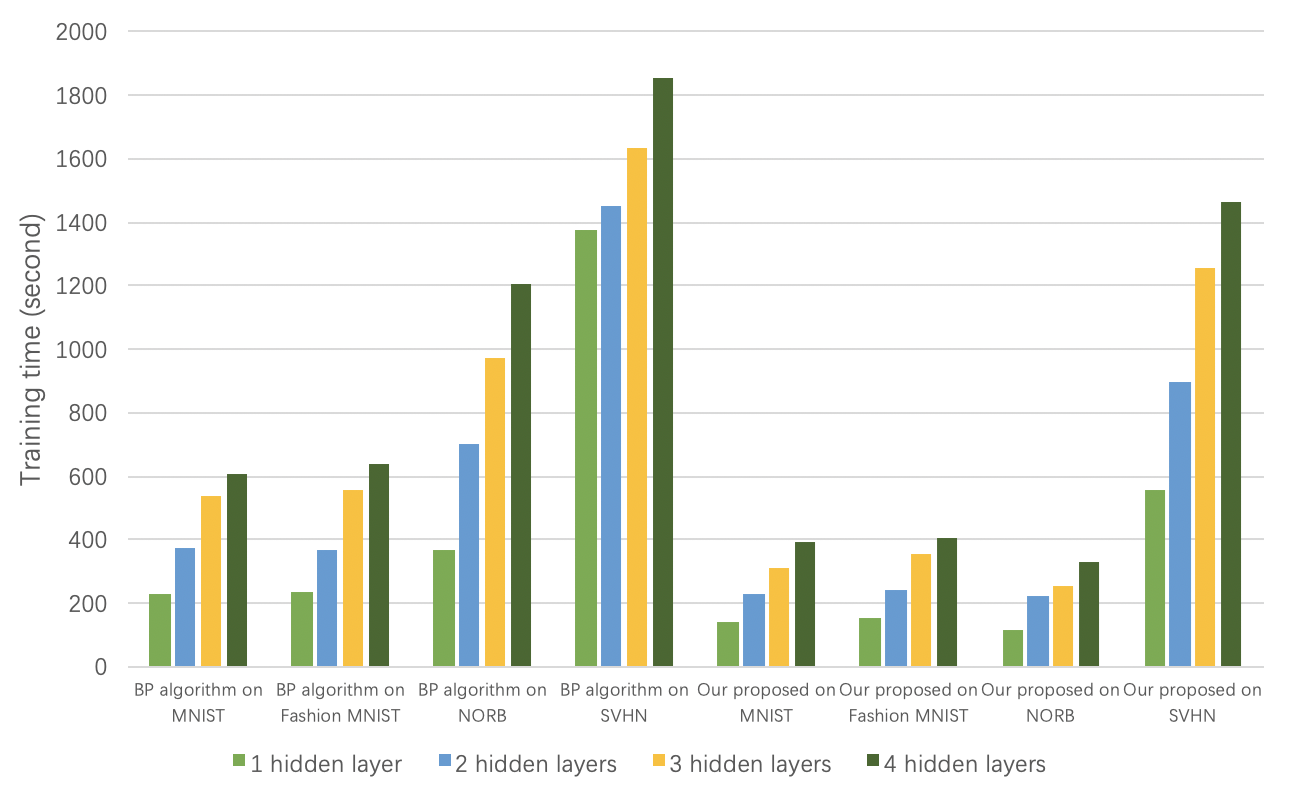}
	\caption{The comparison of training time of the BP algorithm and our proposed method}
	\label{fig_2}
\end{figure}

\subsection{\textit{Discussion}}

\textit{D.1 Advantage}

\textit{D.1.1 The proposed learning scheme is fast}


The proposed algorithm can directly calculate the analytical solution of the optimization objective function by using basic matrix operations, such as matrix product and pseudoinverse. The connection weight matrices are estimated in a feedforward way without any iterative optimization. Thus, it is much faster than the back propagation and other gradient descent-based learning algorithm. This is shown in the above experiments. In real-world applications, not all data are ready before staring training, instead, the data are always collected (or inputted) over certain periods of time due to some limits. In such cases, PILAE could work in an incremental learning way in which the model is updated whenever new samples are available. This version of PILAE is the incremental learning PILAE (IPILAE) which was introduced in our previous work \cite{IEEEhowto:Wang2016}.

\textit{D.1.2 The proposed method is a quasi-automated learning scheme}

There are two types of hyperparameters involving in the stacked autoencoders deep neural network. One is the model hyperparameters which determine the network’s architecture, i.e. the number of the hidden layers (the depth of the network) and the number of hidden neurons in each layer. The other is the control parameter in learning algorithm, such as maximum epoch, step length, weight decay and momentum in gradient based algorithm. These hyperparameters have an important influence on the performance of the stacked autoencoders deep neural network. In the proposed learning scheme, the weight $\mathrm{W}_\mathrm{e}$ connecting the hidden layer and the input layer is estimated by the pseudoinverse of the input matrix with low rank approximation. The weight $\mathrm{W}_\mathrm{d}$ connecting the output layer and the hidden layer is the optimal solution minimizing the reconstruction error function. These connection weight matrices are calculated directly in a feedforward way based on generalized linear algebra theory. The regularization parameters $\lambda_i|_{i=1}^l$ in $i$-th hidden layer is fixed as a very small positive value to guarantee that $\mathrm{H}^+$ in Eq. (\ref{e27}) is positive defined. The regularization parameter $\lambda$ in Eq. (\ref{e12}) can be first estimated with training data as in Eq. (\ref{e3}). And then its optimal estimate is obtained by searching the nearby points around the point estimation. No other control parameters are needed to adjust in the learning process.

Furthermore, the model hyperparameters, that is, the depth of the network and the number of hidden neurons in each hidden layer, are given automatically by the proposed learning scheme. The deep length of the network is determined automatically when the proposed algorithm satisfies the terminal criterion. The hidden layer neuron number in feature learning part is related to input data rank and data dimension, and determined by an empirical formula, Eq. (\ref{e18}) or Eq. (\ref{e19}). And the parameter $\alpha$ or $\beta$ in Eq. (\ref{e18}) or Eq. (\ref{e19}) is estimated based on reconstruction error, that is, with proper reconstruction error setting, we can find a suitable value for $\alpha$ or $\beta$. The neuron number in the classifier could be estimated by the bivariate regression analysis with Eq. (\ref{e30}).

In a word, the proposed learning scheme is not only to improve the learning speed comparing the gradient descent-based learning algorithm, but also to provide an empirical designing of the network structure. Thus, the learning speed of the whole neural network system is improved by the proposed learning scheme. Moreover, it is explored preliminarily to realize the autonomous machine learning (AutoML) of the DNN architecture.

\textit{D.1.3 The proposed learning scheme simplify the hyperparameter setting}

For the layer-wise training of autoencoders, we present an empirical formula to estimate the hidden neuron number. In our learning scheme, searching optimal hyperparameters of DNN structure in multi-dimensional space now is restricted to a search problem in one dimensional space. In our approaches, only two hyper parameters should be optimized: one is the number of neurons in the last hidden layer while another one is the regularization parameter in the loss function. For the first hyper parameter, Eq. (\ref{e30}) can be used to estimate it without exhaust search. For the the second one,  Eq. (\ref{e3}) can be used to speeded up the search in a reduced space. Hence our proposed learning scheme will greatly reduce difficulty of the DNN architecture design.

\textit{D.1.4 The proposed learning scheme is not just a pre-training method}

Pre-training deep neural networks with autoencoders is a basic practice of deep learning. However, the learning scheme we presented in this work is not only a pre-training algorithm, it could be viewed as a non-gradient descent learning algorithm for DNN. Layer-wise training of DNN can be regarded as the application of divide and conquer strategy while an additional fine-tuning is needed in traditional pre-training scheme. However, as Y. Bengio pointed out in Quora.com, the disadvantage of pre-training deep neural networks with autoencoders is that ``\emph{it is greedy, i.e., it does not try to tune the lower layers in a way that will make the work of higher layers easier}''.  To avoid the shortages caused by the greedy learning process, our strategy is to allow small reconstruction error exists in each layer while search the number of hidden neurons in each SHLN to achieve a better generalization performance of DNN.

\textit{D.1.5 The learning model could grow dynamically in a data-driven way}

In the literature, some researchers estimate latent dimension of the input with SVD method, but relationship between latent dimension and the number of hidden neurons in autoencoders has not been investigated before. In fact, what we presented in Section III.A has illustrated that we do not use SVD to estimate latent dimension of the input but use truncated SVD to compute the encoder weight. This is corresponding to the low rank approximation in the field of machine learning. The number of hidden neurons should not be the latent dimension of the input, otherwise only local optima is reached, and it is too greedy to find a good network structure. So we address the DNN network structure optimization problem with stacked autoencoders, which is seldom investigated by other researchers with gradient descent-based learning algorithm. When stacked autoencoders are applied to construct a DNN, we propose a dynamical growth strategy to determine the depth of the DNN, which is quite different with pre-design strategy. In addition, we use an early stop regularization technique to  determine the depth of the network, it can be explained as data driven architecture design.

\textit{D.2 Discussion of Related Work}

(a) \textbf{Related with other non-gradient descent algorithm:} Recently, it is found that under the name of extreme learning machine (ELM), several PIL variants related works have been published \cite{IEEEhowto:Wong, IEEEhowto:Zhang2017, IEEEhowto:Cheng2018}. In fact, ELM is a renamed work, it originates from our work in 1995 \cite{IEEEhowto:Guo1995}, detailed explanations can be found in \cite{IEEEhowto:Guo2019} and more discussions will be presented in the later works.

(b) \textbf{Related with principal component analysis (PCA):} PCA is the optimal linear autoencoder \cite{IEEEhowto:Baldi, IEEEhowto:Bourlard, IEEEhowto:Oja, IEEEhowto:Magdonismail}.  SVD method can be used to calculate pseudoinverse or used in PCA. Bourlard \emph{et al} believed that for autoencoder the nonlinearities of the hidden units are useless and that the optimal parameter values can be derived directly by SVD and low rank matrix approximation. Their approach appears only related linear case, it is a PCA approximation and encoder weight is the same as decoder weight \cite{IEEEhowto:Bourlard}. However, linear network cannot be stacked to form a deep net, because it will collapse into a two layer’s net. While in our proposed method, we apply truncated SVD (low rank approximation) to pseudoinverse of input matrix to compute encoder weight matrix, and then apply nonlinear transformation to obtain hidden layer output.

(c) \textbf{Related with our previous work:} In our previous work \cite{IEEEhowto:Guo2004}, we set hidden neuron number to be equal to the training sample number in order to realize exact learning, while in this work, low rank approximation is adopted for representation learning. This work is the extension of paper \cite{IEEEhowto:Wang2017}, here global optimization of the DNN structure is investigated. We find that Wong \emph{et al}’s work \cite{IEEEhowto:Wong} is more similar with our previous work \cite{IEEEhowto:Guo2004} in learning strategy, the main difference is that activation function is taken as kernel function in their work. Detailed analysis about the similarity and difference will be presented in later work, here we just point out that random projection cannot learn representation from data.

\textit{D.3 Further work}

Currently, the PILAE algorithm can be applied when the loss error obeys Gaussian density distribution, other probabilities such as Laplace probability should be studied in the further work.

For the desired output vector $\mathrm{x}$, the probability model of an autoencoder can be expressed as follows:
\begin{equation}\label{e32}
	\mathrm{x}=\mathrm{W}_\mathrm{d}\mathrm{h}+\epsilon,
\end{equation}
where $\mathrm{h}$ is the output vector of hidden neurons: $\mathrm{h}=f(\mathrm{W}_\mathrm{e}\mathrm{x})$, $\epsilon$ is noise. Suppose $\epsilon\sim\mathrm{N}(0,\sigma_1^2\mathrm{I})$,  we could obtain the following equation with Eq. (\ref{e32}):
\begin{equation}
	p(\mathrm{x}\vert\mathrm{W}_\mathrm{d}\mathrm{h})=\frac{1}{(2\pi)^{d/2}\sigma_1^d}\exp\Big(-\frac{1}{2\sigma_1^2}\parallel\mathrm{x}-\mathrm{W}_\mathrm{d}\mathrm{h}\parallel^2\Big).
\end{equation}

 If the elements in the encoder matrix $\mathrm{W}_\mathrm{d}$ are \emph{i.i.d}, and the prior probability of the element weight $w_j\sim\mathrm{Lapace}(0,\sigma_2)$, that is, $p(w_j)=\frac{1}{2\sigma_2}\exp\big(-\frac{|w_j|}{\sigma_2}\big)$, then the loss function $L$ becomes:
\begin{equation}
	L=\sum_{i=1}^{N}{\parallel\mathrm{x}_i-\mathrm{W}_\mathrm{d}\mathrm{h}\parallel^2}+\frac{\sigma_1^2}{\sigma_2^2}\sum_{j=1}^{dp}\vert w_j\vert.
\end{equation}

This loss function is the sparse coding of an autoencoder, it is also called Lasso in statistics. The proposed PILAE algorithm will be developed for solving the above $L_1$ norm constraint problem in the future.

Here we should note that hidden neuron number and hidden layer number are two hyper-parameters for DNN architecture, it is still an open problem to realize AutoML. Currently, the methods to find an optimal DNN topology architecture include heuristic design, or search in hyper-parameter space. Our work belongs to the later one, but with empirical estimation formula for reducing the region of search space.  This work is our effort toward to AutoML.

Besides, stacked autoencoders based DNN has the drawback for global optima. When we apply sparse or IB constraints to learn data representation, if the proceeding layer output has sparse /low rank, it will become more difficult for succeeding layers to learn more sparse /low rank representation again.  In addition to the strategy to avoid local optima in the proposed learning scheme, other solution is to adopt stacked generalization technique \cite{IEEEhowto:Guo2004} to construct deep and wide learning system for further improving the system performance on given task. Furthermore, some drawbacks of the PILAE may be solved under a general framework named Synergetic Learning Systems we developed recently \cite{IEEEhowto:Guo2020}.

\section{Conclusion}
This paper focuses on developing a non-gradient descent deep learning algorithm, which can reduce the training time for deep feedforward neural network. This is achieved by a feedforward learning algorithm based on pseudoinverse theory and low rank approximation. No iterative optimization is needed for training DNN, and the proposed pseudoinverse learning algorithm with low rank approximation can greatly speed up the learning processing. It is shown by experiments that the proposed learning scheme can reduce the training time drastically. The representation learning performance of the networks with the proposed learning scheme is better than the BP algorithm-based network on the several datasets. The experiments illustrate that our proposed algorithm is a better scheme for training DNN on considering both the time consuming and representation learning capacity.

The limitations of our work could be summarized as follows:

1). Currently, we only explore the MLP based neural network architecture. Nevertheless, it is known that some types of data, such as the image, are not pure data matrices or vectors. The MLP may be not suitable for the extraction of the semantic information in those kinds of data. In the future work, more different system building blocks, such as convolutional layers, and more complicated network structures, such as the residual DNNs and multi-branch DNNs architecture would be explored based on the pseudoinverse learning scheme. Alternatively, we could use other pre-trained deep networks to extract features which are then used as the inputs to the PILAE based network.

2). In this work, only the rank and sample number of the data were used to determine the network architecture. In the future research, more other features of the data should be involved\cite{IEEEhowto:GuoBDDL}.

\ifCLASSOPTIONcaptionsoff
  \newpage
\fi

\end{document}